\documentclass{article}

 \usepackage[dblblindworkshop, final]{neurips_2025}

\usepackage[utf8]{inputenc} 
\usepackage[T1]{fontenc}    
\usepackage{hyperref}       
\usepackage{url}            
\usepackage{booktabs}       
\usepackage{amsfonts}       
\usepackage{nicefrac}       
\usepackage{microtype}      
\usepackage[table]{xcolor}          
\usepackage{graphicx}         
\usepackage{listings}
\usepackage{algorithm}
\usepackage{algorithmic} 
\usepackage{amsmath}
\usepackage{enumitem}

\newcommand{\algorithmname}{PaDA-Agent}

\title{Learning from Generalization Patterns: An Evaluation-Driven Approach to Enhanced
Data Augmentation for Fine-Tuning Small Language Models}

\workshoptitle{Evaluating the Evolving LLM Lifecycle: Benchmarks, Emergent Abilities, and Scaling}

\author{Huan Song \hspace{0.6em}
  {\bf Deeksha Razdan} \hspace{0.6em}
  {\bf Yiyue Qian} \hspace{0.6em}
  {\bf Arijit Ghosh Chowdhury} \\
  {\bf Parth Patwa} \hspace{0.6em}
  {\bf Aman Chadha} \hspace{0.6em}
  {\bf Shinan Zhang} \hspace{0.6em}
  {\bf Sharlina Keshava} \hspace{0.6em}
  {\bf Hannah Marlowe} \\
  \texttt{\{huanso, razdad, iamyiyue, arijitgc,}\\
  \texttt{parthptw, amanchd, shinanz, skeshava, marloweh\}@amazon.com}\\
  AWS Generative AI Innovation Center
  }

\begin{document}

\maketitle

\begin{abstract}
  Small Language Models (SLMs) offer compelling advantages in deployment cost and latency, but their accuracy often lags behind larger models, particularly for complex domain-specific tasks. While supervised fine-tuning can help bridge this performance gap, it requires substantial manual effort in data preparation and iterative optimization. We present \algorithmname~(\underline{Pa}ttern-guided \underline{D}ata \underline{A}ugmentation Agent), an evaluation-driven approach that streamlines the data augmentation process for SLMs through coordinated operations. Unlike state-of-the-art approaches that focus on model training errors only and generating error-correcting samples, \algorithmname~discovers failure patterns from the validation data via evaluations and drafts targeted data augmentation strategies aiming to directly reduce the generalization gap. Our experimental results demonstrate significant improvements over state-of-the-art LLM-based data augmentation approaches for Llama 3.2 1B Instruct model fine-tuning. 
\end{abstract}

\section{Introduction}

Small Language Models (SLMs) \cite{schick2020s,zhang2024tinyllama}\footnote{typically with fewer than 7B parameters} are increasingly attractive for deployment due to their lower cost and latency, but their limited capacity often results in poor generalization on domain-specific tasks. This gap highlights a core evaluation challenge: how do we measure and systematically improve generalization for models that appear well-trained yet fail to transfer beyond the training distribution?

Supervised fine-tuning (SFT) and recent LLM-driven data augmentation methods \cite{dai2025auggpt, lee2024llm2llmboostingllmsnovel, ying2024llms} attempt to address this by expanding training sets with generated examples. However, most approaches emphasize training error correction and overlook the more informative validation failures, where generalization gaps are revealed. Evaluating and exploiting these validation errors is thus critical for understanding how SLMs fall short and for building augmentation strategies that directly target their weaknesses.


In this work, we propose \algorithmname~(\underline{Pa}ttern-guided \underline{D}ata \underline{A}ugmentation Agent), a multi-agent framework that connects evaluation with augmentation for fine-tuning based SLM improvements. \algorithmname~systematically analyzes validation failures to discover error patterns, drafts augmentation strategies, and generates targeted synthetic data with automated quality control. By integrating evaluation into the augmentation loop, our method directly addresses generalization errors rather than treating them as incidental.

Our experiments show that \algorithmname~significantly outperforms state-of-the-art augmentation baselines when fine-tuning the Llama 3.2 1B Instruct model, yielding consistent gains across reasoning, knowledge, and coding benchmarks. Beyond accuracy improvements, the framework produces interpretable augmentation strategies that reveal why models fail -- offering a bridge between evaluation and actionable model improvement.

The primary contributions of this work include:

\begin{itemize}[itemsep=0pt, topsep=0pt]
\item A novel evaluation-guided approach to data augmentation that directly targets generalization gaps by learning from validation errors, and a coordinated multi-agent framework for systematic error analysis and data generation with automated quality control. \looseness=-1
\item Extensive experiments demonstrating consistent generalization improvements across tasks, with an average 6.6-9.2\% performance gain for Llama-3.2-1B-Instruct compared to state-of-the-art data augmentation approaches. \looseness=-1
\end{itemize}

\section{Methodology}

\begin{figure*}[t]
 	\centering
 	\includegraphics[width=\textwidth]{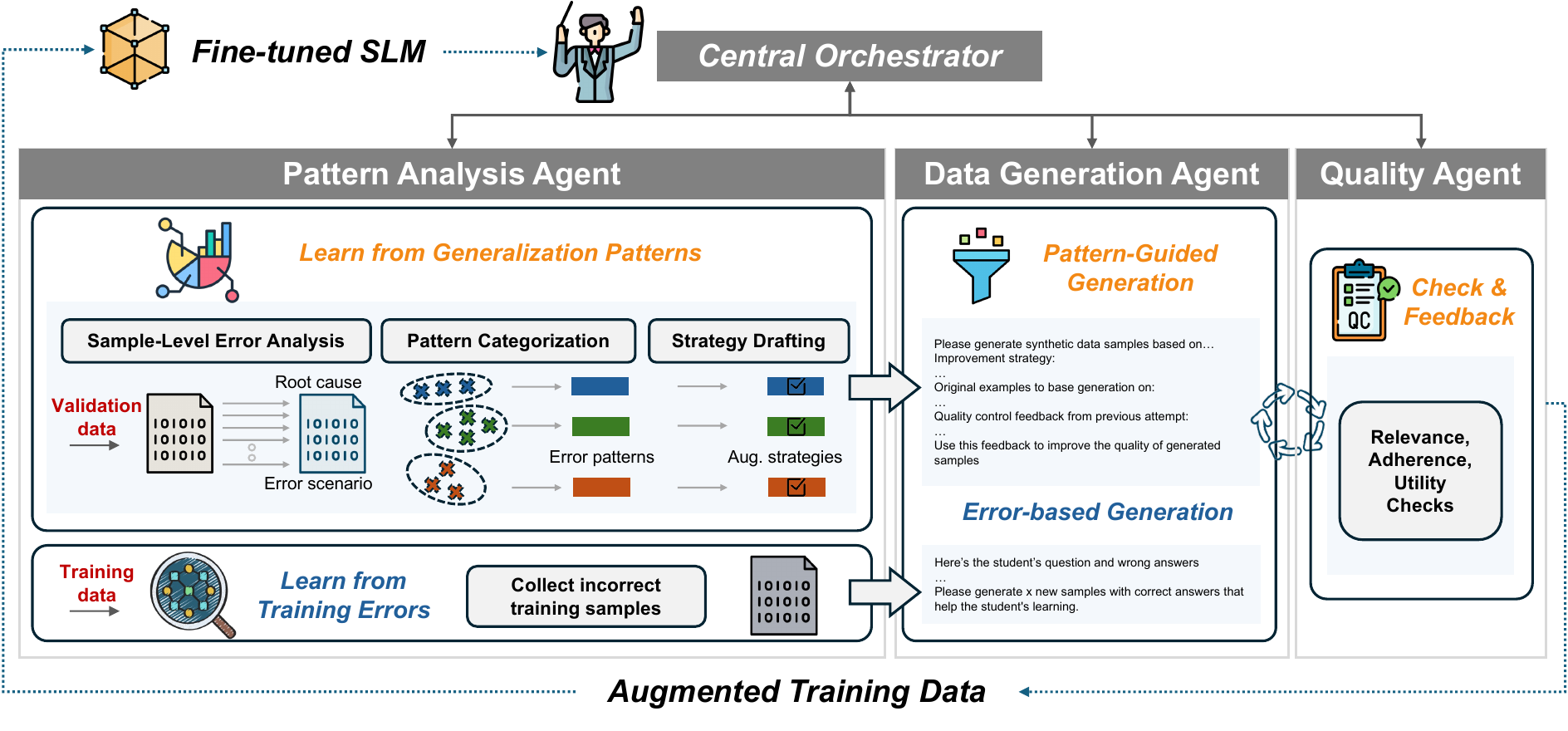}
         	\caption{The architecture of \algorithmname{}: The Central Orchestrator coordinates three specialized agents: the Pattern Analysis Agent (for error analysis, pattern categorization, and augmentation drafting), the Data Generation Agent (for pattern-guided and diverse synthetic data creation), and the Quality Control Agent (for adherence, utility, and relevancy checks). This coordinated process enables improved fine-tuning of SLMs with high-quality, targeted data augmentation.\looseness=-1
        } 
 	\label{fig: agent}
 	\vspace{-0.2in}
 \end{figure*}

\subsection{\algorithmname{}~Architecture}
As shown in Figure~\ref{fig: agent}, \algorithmname{} iteratively augments training data with three agents coordinated by a central orchestrator. The Pattern Analysis Agent extracts generalization failures from validation and errors from training, drafting augmentation strategies. The Data Generation Agent produces synthetic data accordingly, and the Quality Control Agent filters outputs by adherence, utility, and relevancy. Accepted data are added to the training set, the model is re-fine-tuned, and the cycle repeats. The orchestrator maintains a shared state over analyses, strategies, batches, and scores.

Let $\mathcal{D}_{\text{train}}$, $\mathcal{D}_{\text{val}}$, and $\mathcal{D}_{\text{syn}}$ denote training, validation, and synthetic sets. For each $(x_i,y_i)$ with prediction $\hat{y}_i$, failures are
\begin{align*}
\mathcal{E}=\{\,e_i=(x_i,y_i,\hat{y}_i)\mid(x_i,y_i)\in\mathcal{D},\;\texttt{Fail}(\hat{y}_i,y_i,x_i)\,\}.
\end{align*}
We target tasks with verifiable answers (e.g., multiple choice, math, code) and never expose validation samples during training.

\subsection{Pattern Analysis Agent}
Each $e_j\!\in\!\mathcal{E}_{\text{val}}$ is analyzed into $a_j$ (root cause, scenario). Analyses are clustered,
\begin{equation}
\{\mathcal{C}_1,\ldots,\mathcal{C}_K\}=\text{Cluster}(\{a_j\}),
\end{equation}
with $K$ chosen by the elbow method. Each cluster yields a natural-language pattern $\textit{pattern}_k$ and strategy $\textit{strategy}_k$ that guides counterfactual generation.

\subsection{Data Generation Agent}
Pattern-guided samples are generated as
\begin{align*}
\mathcal{D}_{\text{syn}}^{\text{pat}}
=\!\bigcup_{k=1}^{K}\!\{\text{Generate}(x_i,\textit{strategy}_k,\textit{feedback}_i)\mid x_i\in\hat{\mathcal{D}}_{\text{train}}^k\},
\end{align*}
while error-based augmentation corrects training mistakes:
\begin{align*}
\mathcal{D}_{\text{syn}}^{\text{err}}
=\{\text{Generate}(e_i,\textit{feedback}_i)\mid e_i\in\hat{\mathcal{E}}_{\text{train}}\}.
\end{align*}
The final pool is $\mathcal{D}_{\text{syn}}=\mathcal{D}_{\text{syn}}^{\text{pat}}\cup\mathcal{D}_{\text{syn}}^{\text{err}}$.

\subsection{Quality Control and Efficiency}
All synthetic batches are evaluated by the Quality Control agent on adherence, utility, and relevance, each scored on a 1–10 scale. Batches falling below the threshold are regenerated with explicit feedback until quality standards are met. To reduce cost, \algorithmname{} employs batching for generation and evaluation, subsamples validation errors before clustering, and performs pattern analysis at the cluster level, requiring only $K$ calls. Further details on prompts and regeneration heuristics are provided in Appendix~\ref{detailed_method}.

\renewcommand{\arraystretch}{0.8}  

\section{Experimental Results}

\begin{table*}[t]
\centering
\small  
\renewcommand{\arraystretch}{0.9}
\begin{tabular}{lccccccc}
\toprule
Method              & ARC Challenge & GSM8K & HellaSwag & SQuAD & HumanEval & Averaged \\
\midrule
\multicolumn{7}{c}{Standard (1000 train samples)} \\
\midrule
Vanilla Fine-Tuning & 52.6 & 28.3 & 24.2 & 60.4 &  &  \\
AugGPT              & 53.8 & 26.7 & 45.2 & 59.4 &  & +20.4 \\
LLMs-as-Instructors & 49.2 & 27.5 & 47.4 & 63.2 &  & +22.8 \\
Ours                & \textbf{54.6} & \textbf{30.3} & \textbf{51.2} & \textbf{63.6} &  & +32.0 \\
\midrule
\multicolumn{7}{c}{Reduced (600 train samples)} \\
\midrule
Vanilla Fine-Tuning & 50.5 & 26.3 & 21.6 & 59.6 &  &  \\
AugGPT              & \textbf{51.2} & 24.3 & 32.4 & 60.8 &  & +11.5 \\
LLMs-as-Instructors & 50.5 & 28.0 & 35.2 & \textbf{61.2} &  & +18.0 \\
Ours                & 50.5 & \textbf{30.5} & \textbf{40.8} & 60.4 &  & +26.5 \\
\midrule
\multicolumn{7}{c}{Limited (300 train samples)} \\
\midrule
Vanilla Fine-Tuning & 47.3 & 23.4 & 24.6 & 60.6 & 9.4 &  \\
AugGPT              & 45.7 & 18.7 & 27.6 & 59.6 & \textbf{18.8} & -3.1 \\
LLMs-as-Instructors & 50.5 & 27.8 & 28.0 & 60.6 & 12.5 & +9.9 \\
Ours                & \textbf{50.8} & \textbf{28.4} & \textbf{32.6} & \textbf{63.2} & 12.5 & +16.5 \\
\bottomrule
\end{tabular}
\caption{Comparison of Llama 3.2 1B Instruct with vanilla fine-tuning, AugGPT, LLMs-as-Instructors, and \algorithmname~across datasets and data regimes. Numbers are test-set accuracy/EM (\%); best per setting in bold.}
\label{tab:result}
\end{table*}

\subsection{Datasets}
We evaluate across diverse tasks: (1) factual QA with \textbf{SQuAD} v1.1~\cite{rajpurkar2016squad}, (2) commonsense/scientific reasoning with \textbf{ARC Challenge}~\cite{clark2018think} and \textbf{HellaSwag}~\cite{zellers2019hellaswag}, (3) math reasoning with \textbf{GSM8K}~\cite{cobbe2021training}, and (4) coding with \textbf{HumanEval}~\cite{chen2021evaluating}. Metrics are EM for SQuAD, accuracy for ARC/HellaSwag/GSM8K, and pass@1 for HumanEval.

To study low-resource settings, we subsample training data into 1000 (standard), 600 (reduced), and 300 (limited) samples. Each smaller split nests within the larger, ensuring consistency. Validation and test sets (500 samples each, or task-specific) remain fixed across regimes. For HumanEval, we report only the limited setting due to dataset size.



\subsection{Experiment Setup}
For pattern analysis, we subsample 50 validation errors, cluster them (2–10 clusters via elbow method), and draft one strategy per cluster. Data generation produces synthetic samples equal to 50\% of training data, evenly distributed across clusters. Quality control uses a 7/10 threshold with up to three regeneration attempts. All baselines are matched for synthetic data size and iterations.

We fine-tune Llama 3.2 1B Instruct~\cite{grattafiori2024llama3herdmodels} with LoRA ($r=\alpha=32$, dropout=0.05) for 5 epochs at $lr=2e-4$ using Adam. Llama 3.3 70B Instruct powers pattern analysis and generation, while Claude 3.5 Haiku v2 performs quality control to avoid self-enhancement bias. Temperature is 0 for all but data generation (0.7). Training runs on a single NVIDIA A10G.

\subsection{Results}
Table \ref{tab:result} shows that \algorithmname~consistently outperforms vanilla fine-tuning and SOTA baselines. In the 1000-sample setting, it achieves the best results across all tasks, e.g., 51.2\% on HellaSwag vs. 24.2\% baseline, averaging +32.0\%. With 600 samples, gains remain strong (+26.5\%), particularly on GSM8K (30.5\%) and HellaSwag (40.8\%). In the 300-sample regime, \algorithmname~leads on four of five tasks, with notable improvements on ARC (+3.5\%) and GSM8K (+5.0\%). Only HumanEval favors AugGPT.

These results confirm the robustness of our multi-agent approach, especially in low-data regimes, where validation-driven augmentation provides the largest benefit.

\subsection{Ablation Studies}
\label{sec:ablation}
\begin{table}[t]
\centering
\small
\renewcommand{\arraystretch}{0.9}
\begin{tabular}{lcc}
\toprule
\textbf{Method} & \textbf{HellaSwag} & \textbf{ARC Challenge} \\ 
\midrule
\textbf{Full Model (Ours)} & \textbf{39.0} & \textbf{52.6} \\
\midrule
\multicolumn{3}{l}{\textit{Ablation Studies:}} \\
w/o Generalization Patterns & 35.2 (-3.8) & 52.3 (-0.3) \\
w/o Train Errors            & 36.3 (-2.7) & 50.8 (-1.8) \\
w/o Quality Control Agent   & 37.5 (-1.5) & 52.3 (-0.3) \\
\bottomrule
\end{tabular}
\caption{Ablation studies showing the impact of removing different components from our full model. Numbers in parentheses indicate performance drop from the full model.}
\label{tab:ablation}
\end{table}


Table~\ref{tab:ablation} shows that removing generalization pattern analysis causes the largest drop on HellaSwag (-3.8\%), confirming its importance for commonsense reasoning. Eliminating training error analysis also degrades performance (-2.7\% HellaSwag, -1.8\% ARC), while removing quality control yields a smaller decline (-1.5\% HellaSwag) with minimal effect on ARC. These results highlight that both error analysis and quality control contribute meaningfully, with pattern analysis being most critical for generalization.\looseness=-1

\section{Conclusion and Future Work}

We presented a novel multi-agent framework for efficient fine-tuning of SLMs, integrating specialized agents for error pattern analysis, targeted data generation, and quality control. Our experiments demonstrate consistent performance improvements across various tasks, with particularly strong gains in low-data regimes. Our approach underscores the importance of targeted data augmentation and the value of integrating error analysis, generation, and quality control in a cohesive system. Future work should explore the scalability of this approach to larger datasets and models, and investigate pattern transferability across tasks and domains. \looseness=-1


\bibliography{custom}

\begin{thebibliography}{10}

\bibitem{chen2021evaluating}
Mark Chen, Jerry Tworek, Heewoo Jun, Qiming Yuan, Henrique Ponde De~Oliveira Pinto, Jared Kaplan, Harri Edwards, Yuri Burda, Nicholas Joseph, Greg Brockman, et~al.
\newblock Evaluating large language models trained on code.
\newblock {\em arXiv preprint arXiv:2107.03374}, 2021.

\bibitem{clark2018think}
Peter Clark, Isaac Cowhey, Oren Etzioni, Tushar Khot, Ashish Sabharwal, Carissa Schoenick, and Oyvind Tafjord.
\newblock Think you have solved question answering? try arc, the ai2 reasoning challenge.
\newblock {\em arXiv preprint arXiv:1803.05457}, 2018.

\bibitem{cobbe2021training}
Karl Cobbe, Vineet Kosaraju, Mohammad Bavarian, Mark Chen, Heewoo Jun, Lukasz Kaiser, Matthias Plappert, Jerry Tworek, Jacob Hilton, Reiichiro Nakano, et~al.
\newblock Training verifiers to solve math word problems.
\newblock {\em arXiv preprint arXiv:2110.14168}, 2021.

\bibitem{dai2025auggpt}
Haixing Dai, Zhengliang Liu, Wenxiong Liao, Xiaoke Huang, Yihan Cao, Zihao Wu, Lin Zhao, Shaochen Xu, Fang Zeng, Wei Liu, et~al.
\newblock Auggpt: Leveraging chatgpt for text data augmentation.
\newblock {\em IEEE Transactions on Big Data}, 2025.

\bibitem{grattafiori2024llama3herdmodels}
Aaron Grattafiori, Abhimanyu Dubey, Abhinav Jauhri, Abhinav Pandey, Abhishek Kadian, Ahmad Al-Dahle, Aiesha Letman, Akhil Mathur, Alan Schelten, Alex Vaughan, et~al.
\newblock The llama 3 herd of models.
\newblock {\em arXiv preprint arXiv:2407.21783}, 2024.

\bibitem{lee2024llm2llmboostingllmsnovel}
Nicholas Lee, Thanakul Wattanawong, Sehoon Kim, Karttikeya Mangalam, Sheng Shen, Gopala Anumanchipalli, Michael~W. Mahoney, Kurt Keutzer, and Amir Gholami.
\newblock Llm2llm: Boosting llms with novel iterative data enhancement, 2024.

\bibitem{marutho2018determination}
Dhendra Marutho, Sunarna~Hendra Handaka, Ekaprana Wijaya, et~al.
\newblock The determination of cluster number at k-mean using elbow method and purity evaluation on headline news.
\newblock In {\em 2018 international seminar on application for technology of information and communication}, pages 533--538. IEEE, 2018.

\bibitem{rajpurkar2016squad}
Pranav Rajpurkar, Jian Zhang, Konstantin Lopyrev, and Percy Liang.
\newblock Squad: 100,000+ questions for machine comprehension of text.
\newblock {\em arXiv preprint arXiv:1606.05250}, 2016.

\bibitem{reimers2019sentence}
Nils Reimers and Iryna Gurevych.
\newblock Sentence-bert: Sentence embeddings using siamese bert-networks.
\newblock {\em arXiv preprint arXiv:1908.10084}, 2019.

\bibitem{schick2020s}
Timo Schick and Hinrich Sch{\"u}tze.
\newblock It's not just size that matters: Small language models are also few-shot learners.
\newblock {\em arXiv preprint arXiv:2009.07118}, 2020.

\bibitem{ying2024llms}
Jiahao Ying, Mingbao Lin, Yixin Cao, Wei Tang, Bo~Wang, Qianru Sun, Xuan-Jing Huang, and Shuicheng Yan.
\newblock Llms-as-instructors: Learning from errors toward automating model improvement.
\newblock In {\em Findings of the Association for Computational Linguistics: EMNLP 2024}, pages 11185--11208, 2024.

\bibitem{zellers2019hellaswag}
Rowan Zellers, Ari Holtzman, Yonatan Bisk, Ali Farhadi, and Yejin Choi.
\newblock Hellaswag: Can a machine really finish your sentence?
\newblock {\em arXiv preprint arXiv:1905.07830}, 2019.

\bibitem{zhang2024tinyllama}
Peiyuan Zhang, Guangtao Zeng, Tianduo Wang, and Wei Lu.
\newblock Tinyllama: An open-source small language model.
\newblock {\em arXiv preprint arXiv:2401.02385}, 2024.

\end{thebibliography}
\bibliographystyle{plain}

\newpage
\appendix



\section{Detailed metholodgy}
\label{detailed_method}



\subsection{\algorithmname{}~Architecture}
As illustrated in Figure~\ref{fig: agent}, \algorithmname{}~comprise multiple components: pattern analysis, data generation, quality control, all of which coordinated by a central orchestrator. With an initially fine-tuned SLM, pattern analysis agent discovers systematic generalization failures from validation set (learn from generalization patterns), and sample-level mistakes from training set (learn from training errors). It also produces targeted data generation strategies that are passed to the data generation agent. Finally, the quality control Agent rigorously evaluates the generated synthetic data using relevance, adherence, and utility checks, and incorporates data passing the checks into the augmented training dataset. This process operates in an iterative cycle: in the new iteration, the augmented data is used to fine-tune a new version of the SLM, and \algorithmname{}~continues to augment data for the new model. The detailed algorithm for implementing the architecture is shown in Appendix \ref{sec:algorithm}.\looseness=-1

\noindent\textbf{Central Orchestrator.} It manages the overall workflow and state transitions. It first initializes the shared state object and then tracks critical information in the state throughout the workflow, including error analysis results, synthetic data, quality assessments, and configuration parameters etc. \looseness=-1

Next, we provide formal description of each component's implementations in details. We denote the original training set for fine-tuning the SLM $\mathcal{D}_{\text{train}}$, the validation set $\mathcal{D}_{\text{val}}$, and the generated synthetic data $\mathcal{D}_{\text{syn}}$. \algorithmname~requires sample-level evaluation results of these data. While the evaluations can be conducted with LLM-as-a-judge for open-ended tasks, we focus on tasks with verifiable answers such as math reasoning and multi-choice question-answering. For each example $(x_i, y_i)$, the model produces a prediction $\hat{y}_i$. We collect all instances where the model fails on that input as compared to the ground truth:
\begin{align*}
\mathcal{E}
= \left\{ e_i = (x_i, y_i, \hat{y}_i) \,\middle|\,
(x_i, y_i) \in \mathcal{D}, \right. \\
\left. \phantom{= \{} \texttt{Fail}(\hat{y}_i, y_i, x_i) \right\}
\end{align*}
where $\texttt{Fail}$ denotes task-specific failure criterion (e.g. mismatch to ground-truth answer, generated code failing to pass test cases).
\looseness=-1


\subsection{Pattern Analysis Agent}
The pattern analysis agent serves as the analytical foundation of \algorithmname{} by extracting actionable insights that drive targeted data augmentation.\looseness=-1
\subsubsection{Learn from Generalization Patterns}

We propose this pattern-guided augmentation approach based on failure modes made by the model on the validation data. Here, \textit{generalization patterns} refer to systematic error categories that emerge when the model fails on validation data - for example, consistently mishandling multi-step mathematical reasoning, or struggling with specific types of scientific concepts. These patterns, derived by clustering similar validation errors, represent broader model weaknesses rather than isolated mistakes. Note that we derive augmentation strategies from validation error patterns while \textit{never exposing the validation samples themselves to the fine-tuning process}. This design maintains strict separation between training and validation while leveraging validation insights to guide data generation. \looseness=-1





Uncovering the failure modes and design the corresponding augmentation approach is a non-trivial task. We design a sequence of subagents including sample-level error analysis, pattern categorization, and augmentation strategy drafting. \looseness=-1

\noindent\textbf{Sample-Level Error Analysis Subagent.} 
For each error $e_j\in\mathcal{E}_{\text{val}}$, the subagent LLM inspects key features including the user query, model response, ground-truth response, and the evaluation result. It is tasked to determine the potential root cause and the type of scenario where the error occurs (e.g., complex math calculation for math reasoning, historical event recall for knowledge-based question-answering), denoted as $a_j\in\mathcal{A}_{\text{val}}$. The prompt template is shown in Appendix \ref{prompt:error_analysis}. \looseness=-1

\noindent\textbf{Pattern Categorization Subagent.} The pattern categorization subagent synthesizes the sample-level analysis into coherent error patterns in the form of natural language descriptions representing systematic model weaknesses. 
Considering that the error causes could be heterogeneous, we utilize a clustering-based approach to first group the errors. Each error analysis $a_j$ is embedded into a feature vector containing the semantic information. We then apply a clustering algorithm of the features to obtain a grouping of the error analysis into $K$ clusters:\looseness=-1
\begin{equation}
\{\mathcal{C}_1, \ldots, \mathcal{C}_K\} = \text{Cluster}(\{a_j \mid a_j \in \mathcal{A}_{\text{val}}\}).
\end{equation}
We use sentence transformer with all-mpnet-base-v2 embeddings \cite{reimers2019sentence} for the errors, and $k$-means clustering with elbow method \cite{marutho2018determination} that dynamically determines the optimal number of clusters $K$.
Pattern categorization subagent LLM is then applied to each cluster and generate a natural language description $\textit{pattern}_k$ that succinctly summarizes the common characteristics of the errors in that cluster. The prompt template is shown in Appendix \ref{prompt:pattern_categorization}. \looseness=-1

\noindent\textbf{Strategy Drafting Subagent.} Even with the identified error patterns, it could still remain unclear how to address them with data augmentation. This subagent targets this gap by translating identified error patterns into specific strategies for synthetic data generation. For each identified pattern $\textit{pattern}_k$, it develops a corresponding generation strategy $\textit{strategy}_k$ designed to address the weakness. We provide this agent's prompt instruction in Appendix \ref{prompt:strategy_drafting}. A strategy serves as guidance for creating counterfactual examples that demonstrate correct model behavior in similar contexts, thus helping the model learn appropriate responses. Note that each strategy targets a dataset-level error pattern, and is thus readily applicable for data augmentation based on the training set. \looseness=-1


\subsubsection{Learn from Training Errors}
Motivated by existing work that focuses on targeted augmentation of model errors ~\cite{ying2024llms}, we integrate this branch designed to collect incorrect responses of the SLM on the training set $\mathcal{E}_{\text{train}}$, and feed them into the subsequent augmentation agent. The errors made by the model provide valuable learning opportunities for the model to correct itself during the next round of fine-tuning process. 

\subsection{Data Generation Agent}
This agent receives signals from pattern analysis agent for data generation. 
To enable \textit{learn from generalization patterns}, the LLM is tasked to generate samples based on the previously drafted augmentation strategy. Specifically, we randomly sample a training example $x_i\in \mathcal{D}_{\text{train}}$ and generate variants with each augmentation strategy $\textit{strategy}_k$:

\begin{align*}
\mathcal{D}_{\text{syn}}^{\text{pat}} 
= \bigcup_{k=1}^K \big\{ 
\text{Generate}(x_i, \textit{strategy}_k,\, \textit{feedback}_i) 
\,\big|\,
\\
x_i \in \hat{\mathcal{D}}_{\text{train}}^k 
\big\}
\end{align*}
where $\text{Generate}()$ denotes the LLM generation, $\textit{feedback}_i$ denotes the quality improvement feedback provided by the quality control agent (details defined in next section), $\hat{\mathcal{D}}_{\text{train}}^k$ denotes the subsampled training set for strategy $k$. The specific prompt instruction is shown in Appendix \ref{prompt:generation_val}.

To enable \textit{learn from training errors}, the LLM is tasked to generate samples to reinforce the correct learning: 
\begin{equation}
\mathcal{D}_{\text{syn}}^{\text{err}} = \{ \text{Generate}(e_i, \textit{feedback}_i) \mid e_i \in \hat{\mathcal{E}}_{\text{train}} \}, \nonumber
\end{equation}
where $\hat{\mathcal{E}}_{\text{train}}$ denotes subsampled training set for synthetic data generation. The final synthetic dataset is:
\begin{equation}
\mathcal{D}_{\text{syn}} = \mathcal{D}_{\text{syn}}^{\text{err}} \cup \mathcal{D}_{\text{syn}}^{\text{pat}}.
\end{equation}

\subsection{Quality Control Agent}


This agent assesses the generated synthetic data $\mathcal{D}_{\text{syn}}$ in batches and provides a quality score with improvement feedback. 
It assesses the following quality dimensions:\looseness=-1

\begin{itemize}
   \item Adherence to augmentation strategy: How well the example follows the specific data augmentation strategy provided. This only applies to the \textit{learn from generalization patterns} branch.
  \item Training utility: The potential utility of each example for model training.
  \item Relevance to original training sample: Verify whether the synthetic data adheres to the format and style of original training sample. This dimension aims to avoid potential hallucinations.\looseness=-1
\end{itemize}

Each dimension is rated on a 1-10 scale, with specific criteria defining each rating level (detailed instructions given in Appendix \ref{prompt:quality_control}). Per-dimensional scores are averaged into an overall quality score, and compared against a pre-defined threshold. Batches with low scores are returned to the data generation agent for rework (up to a max number of regenerations), together with explicit feedback to guide improvements.\looseness=-1


\subsection{Operational Efficiency}

To improve operation efficiency, we utilize the following approaches: First, we conduct inference-heavy steps in batches. For example, in data generation and quality control agents, the LLM is tasked to generate or evaluate $B$ samples in a single invocation, where $B$ is the batch size. Second, to reduce the inferences required for \textit{learn from generalization patterns} branch, we subsample the validation errors before conducting error analysis and pattern categorization. Note that both pattern categorization subagent and strategy drafting subagent operate on the cluster level and each requires only $K$ LLM invocations. \looseness=-1

\section{Algorithm}
\label{sec:algorithm}
This section shows the algorithm of \algorithmname{}.

\begin{algorithm}[H]
\floatname{algorithm}{Algorithm}
\renewcommand{\algorithmicrequire}{\textbf{Input}}
\renewcommand{\algorithmicensure}{\textbf{Output}}
\caption{\algorithmname~}
\label{alg:muda}
\begin{algorithmic}[1]
\REQUIRE Training set $\mathcal{D}_{\text{train}}$, Validation set $\mathcal{D}_{\text{val}}$, Initial SLM $\mathcal{M}$
\ENSURE Fine-tuned SLM with improved generalization

\STATE \textbf{Initial Fine-Tuning:}
\STATE Fine-tune SLM on $\mathcal{D}_{\text{train}}$

\STATE \textbf{Iterative Improvement:}
\WHILE{not reached max iterations}
    \STATE \textit{// Evaluations}
    \STATE $\mathcal{E}_{\text{train}} \leftarrow$ Evaluate SLM on $\mathcal{D}_{\text{train}}$
    \STATE $\mathcal{E}_{\text{val}} \leftarrow$ Evaluate SLM on $\mathcal{D}_{\text{val}}$    
    \STATE \textit{// Pattern Analysis Agent}
    \STATE$a_j \leftarrow$ ErrorAnalysis($e_j$)
    \STATE$\{\mathcal{C}_1, \ldots, \mathcal{C}_K\} = \text{Cluster}(\{a_j \})$
    \STATE$\textit{pattern}_k\leftarrow$ PatternCategorization($\mathcal{C}_k$)
    \STATE$\textit{strategy}_k\leftarrow$ StrategyDrafting($\textit{pattern}_k$)

    \STATE \textit{// Data Generation Agent}
    \STATE $\mathcal{D}_{\text{syn}}^{\text{pat}} \leftarrow$ Generate($\mathcal{D}_{\text{train}},\{\textit{strategy}_k\}$)
    \STATE $\mathcal{D}_{\text{syn}}^{\text{err}} \leftarrow$ Generate($\mathcal{E}_{\text{train}}$)
    \STATE $\mathcal{D}_{\text{syn}} \leftarrow \mathcal{D}_{\text{syn}}^{\text{err}} \cup \mathcal{D}_{\text{syn}}^{\text{pat}}$

    \WHILE{batch in $\mathcal{D}_{\text{syn}}$}
        \STATE \textit{// Quality Control Agent}
        \IF{QualityScore(batch) < threshold}
            \STATE Regenerate batch with feedback
        \ENDIF
    \ENDWHILE
    
    \STATE \textit{// Model Update Phase}
    \STATE Fine-tune SLM on $\mathcal{D}_{\text{train}} \cup \mathcal{D}_{\text{syn}}$
\ENDWHILE

\RETURN Final fine-tuned SLM
\end{algorithmic}
\end{algorithm}


\section{Prompts}
\label{sec:appendix_pattern_analysis_agent_incorrect}

\begin{figure}[h!]
\begin{lstlisting}[language=Python, breaklines=true, breakindent=0pt]
user_prompt = f"""
Analyze these incorrect responses and identify the root cause and error scenario for each:

SAMPLE 1:
USER QUERY: ...
MODEL RESPONSE: ...
GROUND TRUTH: ...
---
SAMPLE 2:
...
---

Format your response as a JSON array of analysis results. For each analysis, include:
- "sample_idx": <index of sample in batch>,
- "error_cause": "Analysis of what's causing the mistake",
- "scenario_category": "The type of scenario where this error occurs (e.g., 'Complex Math Calculation', 'Historical Event Recall', etc.)"
"""
\end{lstlisting}
\caption{Prompt for Sample-Level Error Analysis Subagent.}
\label{prompt:error_analysis}
\end{figure}

\begin{figure}[h!]
\begin{lstlisting}[language=Python, breaklines=true, breakindent=0pt]
user_prompt = f"""
Please analyze these samples that have been grouped together based on their similarity.
Identify the core error pattern that defines this group of samples.
Focus on the common characteristics and challenges shared by these samples.

ANALYZED SAMPLES IN THIS GROUP:
...

Create a concise categorization that:
1. Identifies the core pattern shared by these samples
2. Describes the specific challenges in this group

Format your response as a JSON object with:
- "category_name": "Descriptive name for this error category",
- "error_pattern": "Core pattern that defines this category of errors",
- "representative_samples": ["1-2 most illustrative samples with their full content"]
"""
\end{lstlisting}
\caption{Prompt for Pattern Categorization Subagent.}
\label{prompt:pattern_categorization}
\end{figure}

\begin{figure}[h!]
\begin{lstlisting}[language=Python, breaklines=true, breakindent=0pt]
user_prompt = f"""
Based on these error pattern categories, generate specific strategies for synthetic data generation.
The objective is to create data generation strategies that will benefit model fine-tuning.
NOTE the strategies and suggestions must be actionable for another large language model for synthetic data generation.
Create one focused strategy per error category that will help the model improve on similar cases. Make sure the strategies are diverse and significantly different.

ERROR CATEGORIES:
...

For each category, create a data augmentation/synthesis strategy that:
1. Directly addresses the identified challenges
2. Provides specific guidance for data generation

Format your response as a JSON array of suggested strategies. For each strategy, include:
- "category_name": "Name of the error category this strategy addresses",
- "strategy_name": "Descriptive name for this strategy",
- "generation_approach": "Detailed approach for generating synthetic data",
- "key_elements": ["Specific elements to include in generated data"].
"""
\end{lstlisting}
\caption{Prompt for Strategy Drafting Subagent.}
\label{prompt:strategy_drafting}
\end{figure}

\begin{figure}[h!]
\begin{lstlisting}[language=Python, breaklines=true, breakindent=0pt]
user_prompt = f"""
Please generate synthetic data samples based on the following improvement strategy and examples.

IMPROVEMENT STRATEGY:
...

ORIGINAL EXAMPLES TO BASE GENERATION ON:
...

QUALITY CONTROL FEEDBACK FROM PREVIOUS ATTEMPT:
...

Use this feedback to improve the quality of generated samples by addressing:
1. Areas where previous samples fell short
2. Specific improvements suggested in the feedback
3. Quality aspects that need enhancement

For each original example, generate {num_samples_per_example} synthetic samples that:
1. IMPORTANT: the synthetic sample will improve fine-tuning of a downstream large language model
2. Follow the strategy's generation_approach
3. Include all key_elements from the strategy
4. Use the example's structure but vary the content such that it differs significantly to ensure diversity
5. Maintain consistency with the strategy's goals
6. Address any quality control feedback if provided

Each synthetic sample should follow the exact same format as the example data.

Format your response as a JSON array of synthetic samples. For each synthetic sample, include:
- "sample_id": A unique identifier (e.g., "synthetic_[random_number]")
- "is_synthetic": true
- "based_on_strategy": "{strategy.get('strategy_name', 'unknown')}"
- "based_on_example": The sample_id of the original example
- "messages": Array with user and assistant messages
"""
\end{lstlisting}
\caption{Prompt for Data Generation Agent - Learn from Generalization Patterns.}
\label{prompt:generation_val}
\end{figure}

\begin{figure}[h!]
\begin{lstlisting}[language=Python, breaklines=true, breakindent=0pt]
user_prompt = f"""
Please generate synthetic data samples based on these training error examples.

EXAMPLE 1:
QUESTION/TASK:
...
STUDENT'S WRONG ANSWER:
...
---
EXAMPLE 2:
...
---

QUALITY CONTROL FEEDBACK FROM PREVIOUS ATTEMPT:
...

Use this feedback to improve the quality of generated samples by addressing:
1. Areas where previous samples fell short
2. Specific improvements suggested in the feedback
3. Quality aspects that need enhancement

Please generate {total_samples} new samples with correct answers that help the student's learning.
Each synthetic sample should follow the exact same format as the original examples but come with the correct response.

Format your response as a JSON array of synthetic samples. For each synthetic sample, include:
- "sample_id": A unique identifier
- "is_synthetic": true
- "messages": Array with user and assistant messages with exactly one-turn of conversation, i.e.
{{"messages":[{{"role":"user","content":<new question>}},{{"role":"assistant","content":<correct response>}}]}}, where <new question> and <correct response> are plain strings without any other structure.
"""
\end{lstlisting}
\caption{Prompt for Data Generation Agent - Learn from Training Errors.}
\label{prompt:generation_train}
\end{figure}

\begin{figure}[h!]
\begin{lstlisting}[language=Python, breaklines=true, breakindent=0pt]
user_prompt = f"""
Please evaluate the quality of synthetic data samples by comparing them with their original counterparts.

IMPROVEMENT STRATEGIES:
...

ORIGINAL-SYNTHETIC PAIRS:
...

For each original-synthetic pair, evaluate and provide specific feedback on:
1. How well the synthetic sample maintains the essential characteristics of the original
2. How effectively it implements the improvement strategies
3. Quality and usefulness for training
4. Specific areas that need improvement
5. What aspects are working well and should be maintained

Rate each synthetic sample on a scale of 1-10, where:
- 1-3: Poor quality, needs major improvements
- 4-6: Moderate quality, needs specific enhancements
- 7-10: High quality, minor or no improvements needed

Format your response as a JSON array of evaluations. Each evaluation object should have:
- sample_id: the ID of the sample
- type: "original" or "synthetic"
- quality_rating: numeric rating from 1-10 (for synthetic samples only)
- feedback: concise feedback on the sample
"""
\end{lstlisting}
\caption{Prompt for Quality Control Agent.}
\label{prompt:quality_control}
\end{figure}

\section{Qualitative Analysis of Generated Patterns: Case Study on ARC-Challenge}

\begin{figure}[t]
\centering
\includegraphics[width=0.6\linewidth]{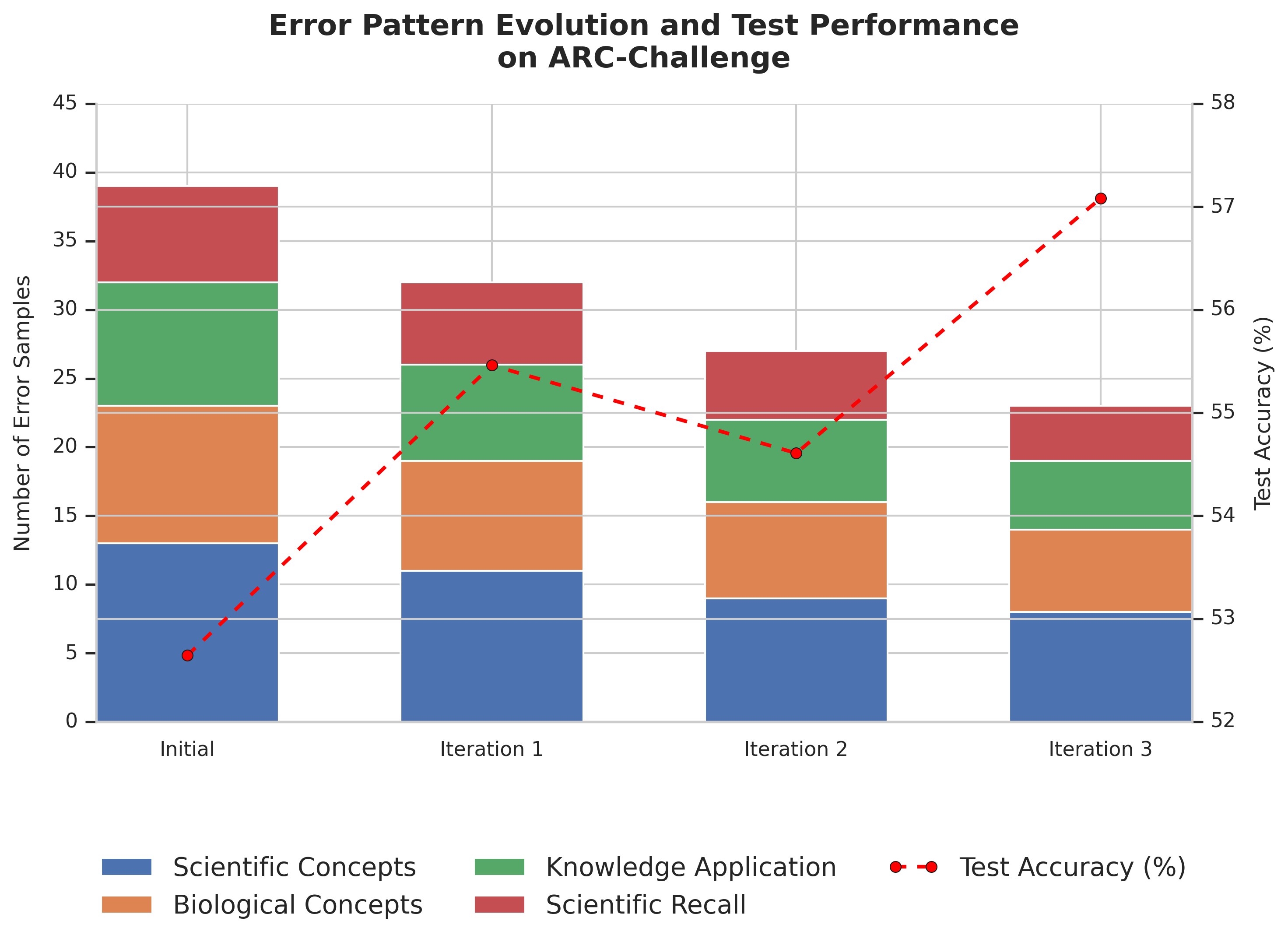}
\caption{Evolution of error patterns across iterations. The decreasing height of stacked bars indicates overall error reduction followed by increase in test accuracy.\looseness=-1}
\label{fig:error_patterns}
\end{figure}

Our qualitative analysis of the ARC-Challenge dataset reveals systematic patterns in model errors and their evolution through iterative fine-tuning. Figure \ref{fig:error_patterns} illustrates the progression of error categories across iterations, demonstrating both the effectiveness of our approach and persistent challenges in specific areas.\looseness=-1

Table \ref{tab:pattern_evolution_complete} provides representative examples of how error patterns evolved from initial to final iterations. Our analysis of all error patterns reveals several insights:

\begin{table*}[t]
\small
\centering
\renewcommand{\arraystretch}{1.2}
\resizebox{0.9\linewidth}{!}{
\begin{tabular}{p{2.6cm}|p{4.8cm}|p{5.8cm}|p{5.2cm}}
\toprule[1.5pt]
\rowcolor{gray!10} \textbf{Category Name} & \textbf{Error Patterns at Iteration 1} & \textbf{Generation Strategy} & \textbf{Error Patterns at Iteration 3} \\
\cmidrule[1pt]{1-4}
Science Knowledge Recall Errors & 
Lack of understanding or inaccurate recall of scientific concepts and principles across various domains, including biology, chemistry, physics, and environmental science  & 
Generate multiple-choice questions that target specific scientific concepts and principles, with a focus on biology, chemistry, physics, and environmental science. Use a mix of easy, medium, and hard questions to challenge the model's recall abilities. & \\
\cmidrule[0.5pt]{1-4}
Biological and Ecological Misconceptions & 
Inadequate understanding of biological and ecological concepts, including evolutionary factors, species relationships, and environmental interactions, leading to incorrect conclusions and answers &  
Create scenarios that test the model's understanding of evolutionary factors, species relationships, and environmental interactions. Use a combination of descriptive text, multiple-choice questions, and case studies to evaluate the model's ability to apply biological and ecological concepts correctly. & Insufficient or inaccurate knowledge of scientific concepts, principles, and processes, particularly in the domains of environmental science, biology, ecology, and conservation, leading to incorrect answers in multiple-choice questions \\
\cmidrule[0.5pt]{1-4}
Scientific Concept Misapplication & 
Misunderstanding or misapplication of scientific concepts, principles, or processes, leading to incorrect conclusions or answers  & 
Create scenarios that require the application of scientific concepts and principles to real-world situations. Use a combination of descriptive text and multiple-choice questions to test the model's ability to apply scientific knowledge correctly. & The core pattern shared by these samples is the misapplication or misunderstanding of fundamental scientific concepts, including physics, thermodynamics, and scientific methodology, leading to incorrect conclusions or answers. \\
\cmidrule[0.5pt]{1-4}
Science and Physics Concept Misapplication & 
Failure to apply fundamental principles and concepts in science, physics, and mathematics, resulting in incorrect conclusions and answers & 
Generate simulated experiments that test the model's understanding of scientific and physics concepts. Use a combination of experimental design, data analysis, and conclusion drawing to evaluate the model's ability to apply scientific principles correctly. & Inability to apply fundamental concepts and principles in various scientific domains, including biology, chemistry, environmental science, and astronomy, resulting in incorrect answers to multiple-choice questions.\\
\bottomrule[1.5pt]
\end{tabular}}
\caption{Example error patterns and generation strategies generated by pattern analysis agent for ARC Challenge dataset.}
\label{tab:pattern_evolution_complete}
\end{table*}

\end{document}